# LL-GaussianMap: Zero-shot Low-Light Image Enhancement via 2D Gaussian Splatting Guided Gain Maps

Yuhan Chen, Ying Fang, Guofa Li, *Senior Member, IEEE,* Wenxuan Yu, Yicui Shi, Jingrui Zhang,
Kefei Qian, Wenbo Chu, Keqiang Li

*Abstract*—Significant progress has been made in low-light image enhancement with respect to visual quality. However, most existing methods primarily operate in the pixel domain or rely on implicit feature representations. As a result, the intrinsic geometric structural priors of images are often neglected. 2D Gaussian Splatting (2DGS) has emerged as a prominent explicit scene representation technique characterized by superior structural fitting capabilities and high rendering efficiency. Despite these advantages, the utilization of 2DGS in low-level vision tasks remains unexplored. To bridge this gap, LL-GaussianMap is proposed as the first unsupervised framework incorporating 2DGS into low-light image enhancement. Distinct from conventional methodologies, the enhancement task is formulated as a gain map generation process guided by 2DGS primitives. The proposed method comprises two primary stages. First, high-fidelity structural reconstruction is executed utilizing 2DGS. Then, data-driven enhancement dictionary coefficients are rendered via the rasterization mechanism of Gaussian splatting through an innovative unified enhancement module. This design effectively incorporates the structural perception capabilities of 2DGS into gain map generation, thereby preserving edges and suppressing artifacts during enhancement. Additionally, the reliance on paired data is circumvented through unsupervised learning. Experimental results demonstrate that LL-GaussianMap achieves superior enhancement performance with an extremely low storage footprint, highlighting the effectiveness of explicit Gaussian representations for image enhancement.

## I. INTRODUCTION

The recovery of high-quality images from degraded visual signals remains a longstanding and challenging fundamental issue in numerous fields including autonomous driving and security surveillance. In this context, increasing attention has been given toward Low-Light Image Enhancement (LLIE). This task aims to alleviate low visibility, low contrast, and severe noise contamination induced by insufficient illumination. High-quality image restoration is essential for accurate visual perception. Furthermore, it serves as a critical prerequisite for the robust performance of high-level vision tasks [1-4].

Substantial advancement in brightness and color enhancement has been achieved in recent years by deep learning-based methods including Convolutional Neural Networks and Transformer architectures. These methods typically learn complex mappings from large-scale datasets. However, images are predominantly treated as discrete pixel grids or are represented through implicit feature embeddings by these methods [5-26]. Consequently, the intrinsic continuity and geometric structure priors of natural images are frequently overlooked in pixel-domain processing. Edge blurring and texture loss are often observed in extremely low-light regions due to the absence of explicit structural guidance. Furthermore, noise tends to be amplified or artifacts tend to be generated during the enhancement process. In addition, most high-performance models rely on strictly paired training data. The acquisition of such data is particularly difficult in real-world scenarios. As a result, the generalization capability of these models is significantly constrained [5-9].

Deep learning paradigms based on unsupervised or semi-supervised learning have received increasing attention to circumvent the reliance on paired datasets [5,7-11,14-15,17-18,20]. However, mapping models in both traditional Retinex-based methods and generative models are formulated in discrete pixel domains. Consequently, continuous signals are not effectively constrained during learning. More importantly, these approaches fail to establish an effective connection between explicit structural representation and illumination enhancement.

2D Gaussian Splatting (2DGS) has recently emerged as an effective explicit scene representation technique in the field of image compression [27-31]. This advantage stems from its strong structural fitting capability and efficient rendering

This work was supported by the National Natural Science Foundation of China under Grant No.52272421. (*Corresponding author: Wenbo Chu*).

Yuhan Chen, Ying Fang, Guofa Li, Wenxuan Yu, Yicui Shi and Kefei Qian are with College of Mechanical and Vehicle Engineering, Chongqing University, Chongqing, 400044, China. (e-mail: 20240701028@stu.cqu.edu.cn; yingfang@stu.cqu.edu.cn; liguofa@cqu.edu.cn; wenxuanyu@cqu.edu.cn; 20212645@cqu.edu.cn; qiankf@stu.cqu.edu.cn).

Wenbo Chu is with the National Innovation Center of Intelligent and Connected Vehicles, Beijing 100089, China (e-mail: chuwenbo@wicv.cn).

Jingrui Zhang is with School of Computer Science, Wuhan University, Wuhan, 430072, China (e-mail: zjr233@whu.edu.cn).

Keqiang Li is with School of Vehicle and Mobility, Tsinghua University, Beijing 100084, China (e-mail: likq@tsinghua.edu.cn).

The code is available at https://github.com/YuhanChen2024/LL-GaussianMap.

mechanism. In contrast to implicit feature embeddings, Image content is explicitly represented by 2DGS as a set of adaptively optimized Gaussian primitives. Each primitive encodes parameters including position, covariance, opacity, and color. Such an explicit formulation enables high-fidelity reconstruction of geometric image structures. Furthermore, high computational efficiency is attained through a tile-based rasterization process. Recent studies have leveraged the strong structural modeling capability of 2DGS for low-level vision tasks [32-35]. However, its direct application to low-light image enhancement has not yet been explored.

To address this limitation, LL-GaussianMap is proposed as the first zero-shot unsupervised framework that introduces 2D Gaussian Splatting into low-light image enhancement. Distinct from existing methodologies, the low-light image enhancement process is reformulated as a gain map generation problem guided by 2DGS primitives. The core procedure of LL-GaussianMap comprises two tightly coupled stages. First, high-fidelity structural reconstruction of the input low-light image is performed using 2DGS, through which intrinsic geometric structure information is accurately captured. Subsequently, a unified enhancement module is developed, where enhancement dictionary coefficients are rendered through the 2DGS rasterization mechanism to generate the final gain map. These coefficients are learned in a data-driven manner and exhibit high compatibility with the reconstructed structure. This two-stage design yields two key advantages. First, the geometric structure explicitly represented by 2DGS is directly exploited to guide gain map generation. Consequently, edge sharpness is effectively preserved, and artifact generation is suppressed during the enhancement process. Second, a zero-shot unsupervised paradigm is adopted for training the entire framework, thereby completely eliminating reliance on paired supervised datasets. As a result, both the generalization capability and practical applicability of the model are substantially enhanced.

The effectiveness of the proposed LL-GaussianMap is evaluated through extensive experiments on multiple public benchmark datasets for low-light image enhancement. The results demonstrate that LL-GaussianMap produces enhanced images with superior visual quality, while effectively retaining fine details and suppressing artifacts. Moreover, its performance exceeds that of numerous state-of-the-art unsupervised methods as well as conventional schemes. More importantly, this study confirms the effectiveness and potential of explicit Gaussian representations for image enhancement, thereby providing new perspectives for future low-level vision studies.

The main contributions of this work are summarized as follows:
- LL-GaussianMap is presented as the first zero-shot unsupervised framework that integrates 2D Gaussian Splatting into low-light image enhancement. The enhancement task is formulated as a gain map generation problem guided by Gaussian primitives.
- A unified enhancement framework is developed by leveraging the explicit structural reconstruction capability of 2DGS. Data-driven enhancement dictionary coefficients are rendered through the Gaussian rasterization mechanism, enabling structure-aware enhancement with effective edge preservation and artifact suppression.
- LL-GaussianMap adopts an unsupervised learning paradigm, thereby eliminating the dependence on paired training datasets. Extensive experimental evaluations further verify the effectiveness of explicit Gaussian representations in image enhancement tasks.

II. RELATED WORKS

LL-GaussianMap lies at the intersection of explicit neural representations and low-level vision tasks, with a particular focus on the use of Gaussian splatting representations for low-light image enhancement. Accordingly, related works are organized into two main categories. First, recent progress in low-light image enhancement is reviewed. Second, existing studies on Gaussian splatting and its representative developments are discussed.

*A. Deep Learning-Based Low-Light Image Enhancement*

Low-Light Image Enhancement (LLIE) aims to improve the visual quality of images captured under low-illumination conditions. It restores brightness, contrast, and fine details while simultaneously suppressing noise and artifacts. This task plays a critical role in various image preprocessing pipelines. Early studies in this field date back to the 1960s, during which traditional methods such as Histogram Equalization and Retinex theory dominated. With the advent of Convolutional Neural Networks, deep learning-based approaches have substantially improved enhancement performance, and have gradually become the mainstream direction in recent research.

Early deep learning-based studies primarily relied on fully supervised learning paradigms, where end-to-end training was performed using strictly paired datasets. For example, Chen et al. proposed FMR-Net and FRR-Net to capture multi-scale features [12-13], and further enhanced feature representation through complex residual network designs. However, fully supervised methods face two major challenges. First, acquiring high-quality paired data is costly, and data collection remains difficult in real-world scenarios. Second, a noticeable domain gap exists between synthetic data and real degraded images, which limits the practical effectiveness of these models.

To reduce dependence on paired datasets, recent studies have increasingly shifted toward unsupervised and zero-shot learning paradigms. Jiang et al. introduced EnlightenGAN, which pioneered the application of Generative Adversarial Networks (GANs) to low-light image enhancement by enabling domain-adaptive enhancement with unpaired data [11]. Subsequently, Guo et al. and Li et al. proposed Zero-Reference Deep Curve Estimation (Zero-DCE) and its improved variant Zero-DCE++ [7-8]. These methods dynamically adjust exposure by iteratively estimating pixel-wise high-order curves, without requiring reference images. Such approaches have substantially advanced unsupervised low-light image enhancement. Building upon this line of work, Pan et al. incorporated Chebyshev approximation theory to improve the robustness of curve estimation [15]. Beyond curve-based methods, recent progress has also been reported in self-supervised learning. Fu et al. learned lightweight enhancement models using paired low-light image instances, and achieved competitive performance [17]. Zhang et al. proposed a noise autoregressive paradigm, which jointly optimizes denoising and enhancement without relying on task-specific data [18]. In addition, Shi et al. presented a zero-shot illumination-guided framework (ZERO-IG) [20], enabling adaptive enhancement and joint denoising in an effective manner.

Purely data-driven methods have demonstrated strong effectiveness, but often lack physical interpretability due to their end-to-end nature. To address this issue, deep unfolding techniques have been developed to integrate physical models with the learning capacity of deep networks. Wu et al. proposed URetinex-Net [19], which unfolds the Retinex decomposition model into a deep network with implicit regularization. Liu et al. designed an unfolding network based on architecture search, allowing atomic priors to be discovered automatically [10]. In addition, Zheng et al. proposed an unfolding-based deep network derived from Total Variation minimization, which enforces fidelity and smoothness constraints and enhances model interpretability and robustness [14].

Lightweight architectures and real-time performance have emerged as critical priorities in resource-constrained scenarios, such as mobile devices and autonomous driving systems. Bai et al. and Chen et al. developed minimalist architectures tailored for mobile and embedded platforms [5,21], with fewer than 200 learnable parameters in a single model. Moreover, Ma et al. introduced the Self-Calibrated Illumination (SCI) framework [9], which employs cascaded illumination learning and weight sharing. A self-calibration module is further incorporated to accelerate convergence, enabling efficient, flexible, and robust image enhancement.

In recent years, generative diffusion models and implicit neural representations have brought renewed interest to low-light image enhancement (LLIE). LightenDiffusion and Aglldiff leverage the strong generative priors of diffusion models to enable unsupervised high-quality image reconstruction [22-23], thereby alleviating over-smoothing and texture degradation. At the same time, Yang et al. and Chobola et al. investigated the use of implicit neural representations for image enhancement tasks. By modeling images as continuous functions, these approaches achieve resolution-agnostic and context-aware enhancement, offering an alternative solution for processing low-light images at arbitrary resolutions [25-26].

*B. Gaussian Splatting*

3D Gaussian Splatting (3DGS) is widely regarded as a milestone in explicit radiance field representation, driving substantial progress in novel view synthesis and 3D reconstruction in recent years [27]. Unlike Neural Radiance Fields (NeRF), which rely on implicit coordinate-based mappings, 3DGS represents scenes using anisotropic three-dimensional Gaussian ellipsoids as explicit geometric primitives. Differentiable tile-based rasterization is further incorporated, enabling real-time rendering while preserving high-fidelity visual quality. This explicit and efficient representation has attracted considerable attention, leading to extensions across a wide range of complex scene reconstruction tasks.

In autonomous driving and dynamic scene modeling, Gaussian primitives have been employed to represent dynamic urban environments and moving vehicles. Through the integration of world models, several studies have explored the generation and reconstruction of four-dimensional driving scenes [36,41-42,46-47]. To address challenges associated with large-scale environments and reconstruction efficiency, multiple extensions of 3DGS have been proposed. By introducing multi-view stereo geometric priors, momentum-based self-distillation, and block-level parallel rendering strategies, 3DGS has been successfully scaled to city-level scenarios [38,44-45], resulting in reduced memory consumption and improved computational efficiency.

Other studies focus on accelerating reconstruction under sparse input conditions, with training efficiency further improved through progressive propagation strategies [37,43]. Beyond reconstruction, the application scope of 3DGS has expanded to text-to-3D content generation, spatiotemporal ballistic motion reconstruction, and scene enhancement under low-light conditions. These efforts collectively demonstrate the versatility and robustness of 3DGS in handling complex three-dimensional data [35,39-40,48].

Motivated by the strong capability of 3D Gaussian Splatting (3DGS) in modeling complex geometry and textures, recent studies have explored its dimensionality reduction for two-dimensional image representation tasks, with the aim of replacing traditional implicit neural representations. The 2D Gaussian Splatting (2DGS) paradigm was first introduced by GaussianImage [30], which represents images as collections of two-dimensional Gaussian primitives. This formulation enables ultra-fast image encoding and decoding speeds exceeding 1000 FPS together with high compression ratios, while demonstrating the potential to outperform coordinate-based neural networks. Subsequent works further investigated the performance limits of 2DGS across different image resolutions and representation scales. Instant GaussianImage proposed an adaptive representation framework with improved generalization capability [29]. LIG focuses on high-resolution image modeling and introduces a hierarchical 2DGS strategy, effectively addressing fine-grained representation for large-scale images [28]. Beyond pure image fitting, the efficient parameterization of 2DGS has been extended to dataset distillation, where sparse Gaussian representations substantially reduce data storage requirements [31]. Compressed 2DGS-based image representations have also been employed to improve the efficiency of multimodal models in vision–language alignment tasks [32]. Collectively, these studies establish 2DGS as an efficient, compact, and differentiable image representation for computational photography and fundamental computer vision applications.

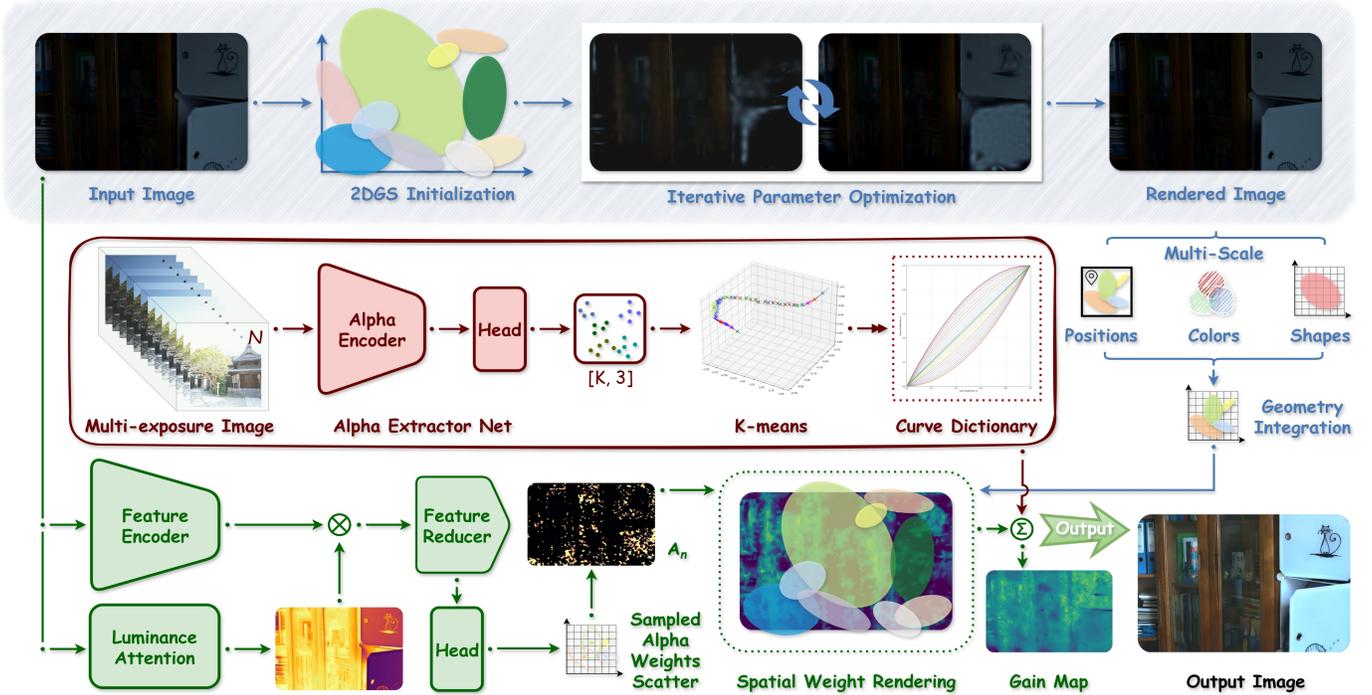

**Fig. 1.** Overall architecture of the proposed LL-GaussianMap framework. The framework introduces a novel enhancement paradigm that integrates explicit geometric representation with data-driven manifold priors. The pipeline consists of three main stages: 1) the input low-light image is represented as a set of discrete 2D Gaussian primitives, whose geometric parameters are subsequently frozen as spatial anchors; 2) a compact enhancement dictionary composed of typical illumination transformation atoms is constructed offline from large-scale data; 3) a lightweight network predicts dictionary mixing coefficients for each Gaussian primitive, and a structure-aware continuous gain map is generated through differentiable rasterization of the weighted atoms. The final enhanced image is obtained by applying the gain map to the input image via pixel-wise multiplication.

Inspired by the success of GaussianImage, the continuity and differentiability of Gaussian functions have been increasingly leveraged to address low-level vision problems. Unlike conventional CNN-based or Transformer-based approaches, 2DGS-based methods model image signals as explicitly parameterized continuous functions. In the field of image super-resolution, GaussianSR and Pixel-to-Gaussian pioneer the use of 2DGS to achieve arbitrary-scale reconstruction [33-34]. These methods learn mappings from image pixels to Gaussian distributions, thereby avoiding artifacts introduced by traditional interpolation schemes. Benefiting from the high degree of parallelism in Gaussian rasterization, a favorable trade-off between inference efficiency and reconstruction fidelity is achieved. These findings indicate that 2DGS extends beyond a static image storage format, acting as a dynamic computational primitive with powerful generative and restorative capabilities, and providing novel solution paradigms for image restoration and enhancement.

## III. PROPOSED METHOD

The proposed LL-GaussianMap framework is described in this section, with its overall architecture shown in Fig.1. The framework consists of three main components. First, multi-scale explicit 2D Gaussian radiance fields are constructed to capture fine-grained geometric structures within the scene. Second, illumination enhancement primitives are learned from large-scale datasets in an unsupervised manner, forming a data-driven manifold-based enhancement dictionary. Finally, a structure-aware unified enhancement module is proposed, in which frozen Gaussian parameters are used as geometric priors to adaptively fuse dictionary atoms and render them across the spatial domain, achieving precise pixel-level enhancement.

### A. Explicit 2D Gaussian Radiance Fields Construction

To obtain accurate geometric and texture representations for subsequent enhancement, the input low-light image $I_{low} \in \Re^{H \times W \times 3}$ is explicitly reconstructed using a set of discrete 2D Gaussian Splatting (2DGS) primitives. Unlike implicit neural representations (INR), 2DGS enables direct modeling and manipulation of geometric attributes in the image space. The image plane is represented as a continuous radiance field composed of individual two-dimensional Gaussian kernels. Each Gaussian primitive $\mathbf{g}_i$ is parameterized by a center position $\mu_i \in \Re^2$, covariance matrix $\Sigma_i \in \Re^{2 \times 2}$, and color coefficients $c_i \in \Re^3$. For a pixel location $x \in \Re^2$, the response value of the $i$ Gaussian primitive $\mathbf{g}_i$ is defined as:

$$G_i(x) = \exp\left(-\frac{1}{2}(x - \mu_i)^T \Sigma_i^{-1} (x - \mu_i)\right). \quad (1)$$

To ensure the semi-positive definiteness and physical interpretability of the covariance matrix $\Sigma_i$, it is decomposed into the product of a scaling matrix and a rotation matrix:

$$\sum_i = R_i S_i S_i^T R_i^T$$
$$= \begin{bmatrix} \cos\theta & -\sin\theta \\ \sin\theta & \cos\theta \end{bmatrix} \begin{bmatrix} s_x & 0 \\ 0 & s_y \end{bmatrix} \begin{bmatrix} s_x & 0 \\ 0 & s_y \end{bmatrix}^T \begin{bmatrix} \cos\theta & -\sin\theta \\ \sin\theta & \cos\theta \end{bmatrix}^T, \quad (2)$$

where $\theta$ denotes the rotation angle, and $S_x$, $S_y$ denote the scaling factors along the two principal axes. For image rendering, a sort-based volumetric rendering approximation is employed, commonly referred to as the splatting process. For an arbitrary pixel location x, the reconstructed color $\hat{I}(x)$ is obtained by accumulating contributions along the depth order:

$$\hat{I}(x) = \sum_{i \in Q} c_i \alpha_i G_i(x) \prod_{j=1}^{i-1}(1 - \alpha_j G_j(x)), \quad (3)$$

where $\alpha_i$ denotes the opacity coefficient, and Q represents the set of Gaussian primitives covering pixel $x$, sorted by depth. To capture full-spectrum information spanning from low-frequency illumination to high-frequency textures, the direct optimization of a single-resolution Gaussian field is avoided. Instead, an efficient multi-scale reconstruction paradigm inspired by the LIG framework is adopted [28]. The number of pyramid levels is denoted by $S$, and Gaussian primitives are hierarchically optimized at resolutions $\{(H_s, W_s)\}_{s=0}^{S-1}$. The reconstruction target at scale $S$, denoted as $I_{target}^{(s)}$, is defined as the residual from the previous level:

$$I_{target}^{(s)} = \begin{cases} DownSample_\downarrow(I) & \text{if } s = 0, \\ I_{ds}^{(s)} - UpSample_\uparrow(\hat{I}^{(s-1)}) & \text{if } s > 0. \end{cases} \quad (4)$$

where $\hat{I}^{(s-1)}$ represents the cumulative reconstruction obtained from the preceding $S - 1$ levels. At scale $S$, the Gaussian parameter set $\Theta_s = \{\mu, \Sigma, c\}_s$ is optimized by minimizing the following photometric loss:

$$\ell_{rec}^{(s)} = (1 - \lambda) \|\hat{I}^{(s)} - I_{target}^{(s)}\|_1 + \lambda\left(1 - SSIM(\hat{I}^{(s)}, I_{target}^{(s)})\right). \quad (5)$$

The final reconstructed image $I_{rec}$ is obtained via the cascaded superposition of rendering results from all scales:

$$I_{rec} = \sum_{s=0}^{S-1} UpSample_{\uparrow \to 0}(Rasterize(\Theta_s)). \quad (6)$$

Beyond achieving high-quality image reconstruction, this stage crucially yields a set of frozen Gaussian parameters denoted as $\theta_{frozen} = \bigcup_s \Theta_s$. These parameters provide precise representations of image edges, textures, and geometric structures, and are subsequently employed as structural priors in the enhancement stage, thereby guiding the spatial distribution of enhancement weights.

*B. Data-Driven Manifold Enhancement Dictionary Learning*

Traditional image enhancement methods often rely on fixed mathematical models such as Retinex and Histogram Equalization or completely black-box end-to-end networks. The former lacks flexibility, whereas the latter suffers from a deficit in interpretability. To integrate the advantages of both, a data-driven manifold enhancement dictionary is constructed on a large-scale unsupervised dataset. Referring to the dataset construction of ZeroDCE, the assumption is made that complex non-linear illumination transformations can be decomposed into a linear combination of a set of basic transformation operators [7-8]. First, a parameterized pixel-level transformation function $T(v; a)$ is defined, where $v$ denotes the pixel value and $a \in \mathcal{R}^P$ represents the parameter vector. Inspired by classic image processing, a quadratic curve model is adopted to fit illumination adjustments:

$$T(v, a) = v + \sum_{p=1}^{P} a_p (v^2 - v). \quad (7)$$

This formula simulates different degrees of exposure gain in the simplified case of $P = 1$. As illustrated in Fig.1, a lightweight feature extraction network, Alpha Extractor Net, is designed to learn these parameters from data. This network is denoted as $\mathcal{E}_\phi$. Images I are taken as input and Global parameter vectors $\hat{a} = \varepsilon_\phi(I)$ are output. The training of the network is aimed at making the brightness of the enhanced image approximate the target value $E_{ref}$.

$$\ell_{dict} = \|mean\left(T(I; \varepsilon_\phi(I))\right) - E_{ref}\|_2^2. \quad (8)$$

Upon completion of the training phase, the network $\mathcal{E}_\phi$ is executed on the entire dataset and a massive collection of parameter vectors $A_{total} = \{\hat{a}_k\}_{k=1}^{Data}$ is acquired. These vectors constitute the manifold distribution of the enhancement space. As illustrated in Fig.2, the K-Means clustering algorithm is performed on $A_{total}$ to construct a compact dictionary further. The following optimization problem is solved to obtain K cluster centers $\{c_j\}_{j=1}^K$:

$$\min_{\{c_j\}} \sum_{k=1}^{|Data|} \min_{j \in \{1,\dots,K\}} \|\hat{a}_k - c_j\|_2^2, \quad (9)$$

where the cluster centers $C_j \in \mathcal{R}^P$ function as the dictionary atoms. To reinforce the identity mapping capability of the model, a zero vector $c_0 = 0$ is explicitly incorporated into the dictionary, signifying the null operation. The resulting enhancement dictionary matrix $D \in \mathcal{R}^{(K+1) \times P}$ is defined as follows:

$$D = [c_0, c_1, \dots, c_K]^T. \quad (10)$$

As depicted in Fig.3, a fundamental illumination transformation pattern characterized by data statistical significance is represented by $D_k$ in the k row.

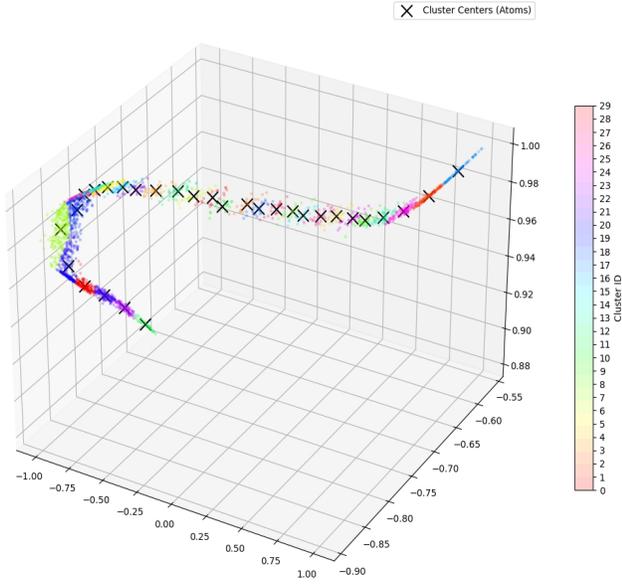

**Fig. 2.** Visualization of the learned enhancement manifold in parameter space. It is verified by this visualization that a compact low-dimensional manifold is formed by the enhancement priors of natural images. Furthermore, this manifold is effectively spanned by a set of discrete basis vectors.

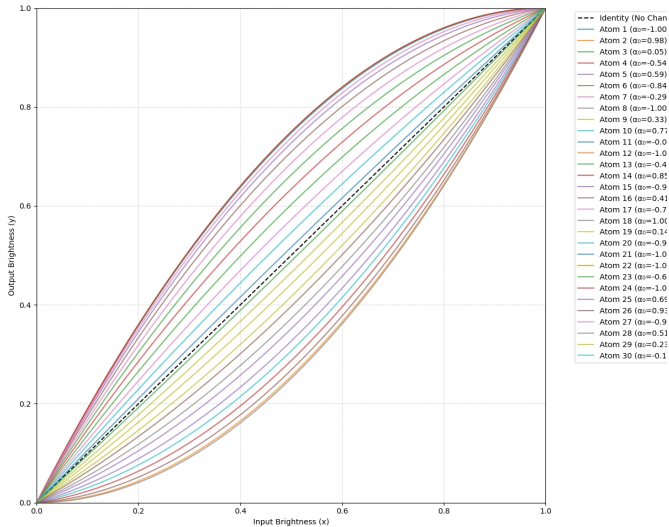

**Fig. 3.** Basis transformation curves constituting the manifold enhancement dictionary. It is verified by this visualization that a compact low-dimensional manifold is formed by the enhancement priors of natural images. Furthermore, this manifold is effectively spanned by a set of discrete basis vectors. Each curve corresponds to a specific atom vector within the dictionary matrix.

### C. Structure-Aware Gaussian Image Enhancement

The implementation of low-light image enhancement utilizing the reconstructed explicit 2DGS geometric structure $\Theta_{frozen}$ and the acquired enhancement dictionary D is detailed in this section. The fundamental philosophy of LL-GaussianMap dictates that the spatial distribution of enhancement coefficients must be aligned strictly with the physical structure of the image. To this end, the intrinsic rendering capabilities of 2DGS are leveraged to generate the enhancement map, rather than relying on traditional bilinear interpolation.

**Gaussian-guided coefficient inference and rasterization.** A lightweight Convolutional Neural Network $H_\varphi$ is designed to predict a spatially varying dictionary coefficient index map, given an input low-light image $I_{in}$. To strike a balance between computational efficiency and feature representation capability, a compact encoder-decoder architecture based on a pre-trained MobileNetV2 is constructed, rather than employing the cumbersome standard U-Net. Specifically, the first seven stages of the pre-trained MobileNetV2 are extracted to serve as the feature extraction backbone. The input image $I_{in}$ is mapped into a deep semantic feature tensor $F_{enc} \in \Re^{C_b \times H' \times W'}$ by this backbone network. Considering the unique nature of low-light enhancement tasks where dark regions necessitate more pronounced adjustments than bright areas, a brightness-guided attention mechanism is introduced to modulate encoder features. As illustrated in Fig.4, the grayscale brightness map $L \in \Re^{1 \times H \times W}$ of the input image is calculated first and an inverted attention mask $M_{att} = 1 - L$ is constructed to emphasize low-illumination regions. This mask is resized via bilinear interpolation to match the spatial resolution of the features $F_{enc}$. Subsequently, element-wise multiplication is performed between the attention mask and the encoded features. Thus, attention-modulated features $F_{att}$ are obtained:

$$F_{att} = F_{enc} \otimes \text{Re size}_{\downarrow 8}(M_{att}), \quad (11)$$

Where $\otimes$ denotes element-wise multiplication. As illustrated in Fig.4, feature responses in well-exposed regions are suppressed effectively by this step, thereby directing the network to focus on under-exposed details necessitating restoration. Finally, a shallow Decoding Head is devised to project features into the coefficient space. This component is composed of two consecutive convolutional layers.

To map these low-frequency weights to the high-resolution pixel space while maintaining edge sharpness, sampling $W_{low}$ is performed utilizing the Gaussian positions $\mu_i$ from the reconstructed explicit 2DGS geometric structure.

For the $i$ Gaussian point, its corresponding dictionary mixing weight vector $w_i$ is expressed as:

$$w_i = GridSample(W_{low}, \mu_i), \quad w_i \in \Re^{K+1}, \quad (12)$$

where $\mu_i$ is normalized to the interval $[-1,1]$. Subsequently, a critical Coefficient Splatting is executed. By treating the weight vector $w_i$ of each Gaussian primitive as a color attribute, the rendering process is performed via the Gaussian rasterizer utilizing frozen geometric parameters $\Theta_{frozen}$. The resulting full-resolution weight map $\Omega \in \Re^{(K+1) \times H \times W}$ is represented as follows:

$$\Omega(x) = \sum_{i \in Q} Softmax(w_i) \cdot \alpha_i G_i(x) \prod_{j=1}^{i-1}\left(1 - \alpha_j G_j(x)\right), \quad (13)$$

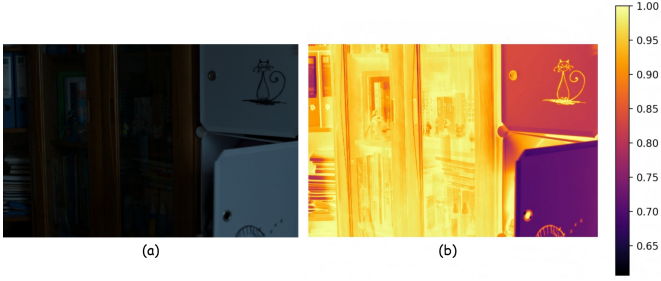

**Fig. 4.** Visualization of the brightness-guided attention mechanism. (a) Input low-light image. (b) Generated attention map $M_{att}$. Darker regions in the input image correspond to brighter regions in the attention map. Consequently, the lightweight encoder is guided effectively to prioritize feature extraction in underexposed regions. Simultaneously, well-exposed backgrounds are suppressed.

where Softmax is utilized to ensure that the sum of weights at each pixel position equals 1. The geometric edge information of the original image is inherited naturally by the weight map rendered in this manner. Consequently, the structure-aware characteristic is realized. This capability is unmatched by traditional CNN upsampling.

**Gain map construction and image generation.** Upon acquisition of the pixel-level weight map $\Omega(\mathcal{X})$, the final illumination gain map $\eta(x)$ is constructed by linearly combining dictionary atoms. As illustrated in Fig.5, let $\Omega_k(x)$ be the value of the k channel of $\Omega_k(x)$. This corresponds to the dictionary atom $D_k$. The parameter $\gamma(x)$ of the gain map is defined as:

$$\gamma(x) = \sum_{k=0}^{K} \Omega_k(x) \cdot D_k. \tag{14}$$

Based on the quadratic transformation model defined in the (7), the spatially varying gain map $\eta(x)$ is obtained and depicted in Fig. 6. The formula is expressed as:

$$\eta(x) = 1 + \gamma(x). \tag{15}$$

Additionally, a bias term $b \in \mathfrak{R}^3$ predicted by global features is introduced to address global color deviation and black level offset. The final enhanced image $I_{enhance}$ is represented as:

$$I_{out}(x) = Clamp(I_{low}(x) \otimes \eta(x) + b). \tag{16}$$

**Hybrid loss function optimization.** A comprehensive set of loss functions is designed to train the enhancement network $H_\varphi$ effectively. Exposure control, spatial consistency, sparsity constraints, and perceptual quality are encompassed by this objective. First, global uniform brightness is not enforced to address non-uniform illumination. Instead, local targets are constructed based on Retinex theory. A local adaptive target loss $\ell_{target}$ is introduced:

$$\ell_{target} = \|I_{out} - I_{gt}\|_1, \tag{17}$$

where the target brightness is constructed by calculating the input image brightness $L_{low}$ and its Gaussian blurred version $L_{blur}$ as follows:

$$I_{gt} = Clamp\left(I_{low} \cdot \frac{E_{target}}{I_{blur} + \varepsilon}\right). \tag{18}$$

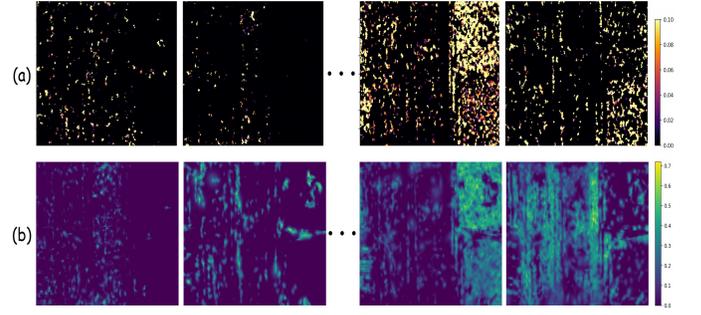

**Fig. 5.** Propagation from sparse prediction to dense structure-aware weight maps. (a) Scattered dictionary coefficients $w_i$ sampled at discrete Gaussian centers are represented. The adaptive density of the explicit Gaussian representation is reflected by the sparsity of points. (b) The corresponding dense weight map $\Omega_k$ obtained via Gaussian splatting is displayed. The structure-aware characteristic of Gaussian-guided rendering is verified by this result.

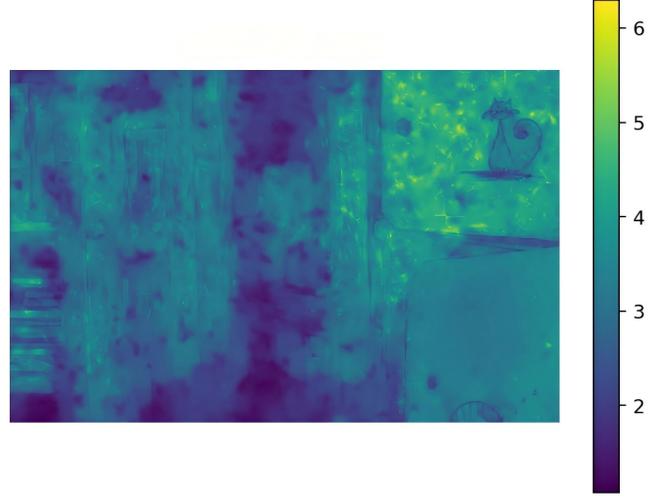

**Fig. 6.** Visualization of the final spatial gain map $\eta(x)$ synthesized from dictionary atoms. Smooth variations are exhibited within homogeneous regions. Simultaneously, sharp transitions are maintained at structural boundaries. Local adaptive enhancement is realized by this spatially varying gain field. Details in bright regions are preserved while dark regions are enhanced significantly.

Simultaneously, consistency between the enhanced map and the original image in the gradient domain is constrained to preserve enhanced texture details. Consequently, a spatial consistency loss $\ell_{spa}$ is introduced, where $\nabla_x, \nabla_y$ is defined as the gradient operator:

$$\ell_{spa} = \sum_{d\in\{x,y\}} \|\nabla_d I_{enhance} - \nabla_d I_{low}\|_1. \tag{19}$$

To suppress over-exposure and under-exposure, the local mean value of the image is constrained. Thus, an exposure consistency loss $\ell\exp$ is introduced. The image is partitioned into $Z_y$ patches $Y_j$ of size $16 \times 16$. Subsequently, the average intensity $\overline{Y}_j$ is calculated:

$$\ell_{\exp} = \frac{1}{Z_y} \sum_{j=1}^{Z_y} \left\|\overline{Y}_j - E_{t\,arg\,et}\right\|_2^2. \tag{20}$$

To guarantee the conciseness of the solution and prevent excessive mixing of dictionary atoms, a dictionary sparsity loss $\ell_{sparse}$ is introduced. An L1 sparsity constraint is imposed on the weights $W_{sample}$ prior to rendering. This constraint is specifically targeted at non-zero atoms and is expressed as:

$$\ell_{sparse} = E[\|w_{i,1:K}\|_1]. \quad (21)$$

Simultaneously, a Total Variation constraint $\ell_{tv}$ is imposed on the gain map $\eta(x)$ to guarantee the smoothness of illumination variations and prevent artifact generation:

$$\ell_{tv} = \|\nabla_x \eta\|_1 + \|\nabla_y \eta\|_1. \quad (22)$$

Finally, a perceptual contrast loss $\ell_{cont}$ is introduced to elevate visual clarity. The gradient magnitude of the enhanced image is encouraged to remain not inferior to that of the original image:

$$\ell_{cont} = \text{Relu}(mean(M_{low}) - mean(M_{enhance})). \quad (23)$$

The final total loss function is defined as the weighted sum of the aforementioned terms:

$$\begin{aligned}\ell_{total} &= \lambda_1 \ell_{target} + \lambda_2 \ell_{spa} + \lambda_3 \ell_{exp} \\ &+ \lambda_4 \ell_{sparse} + \lambda_5 \ell_{tv} + \lambda_6 \ell_{cont.}\end{aligned} \quad (24)$$

Natural and robust illumination enhancement is achieved via data-driven dictionaries through the joint optimization of the aforementioned objectives. Simultaneously, image details are preserved effectively utilizing explicit geometric structures.

IV. EXPERIMENTS AND RESULTS

*A. Experimental Setup*

**Dataset.** Two benchmark datasets are utilized to assess the performance of the LL-GaussianMap model. Specifically, the LOL dataset and the Large-Scale Real-World (LSRW) dataset are selected [49-50]. LOL is established as the premier public paired dataset meticulously constructed for supervised learning tasks in low-light image enhancement. The data comprises synthetically generated low-light images alongside corresponding real-world normal-light images. As the inaugural large-scale real-world paired dataset for low-light and normal-light imagery, two independent subsets are contained within LSRW. Images were acquired via a Huawei P40 Pro smartphone and a Nikon D7500 digital SLR camera respectively.

**Implementation Details.** The proposed structure-aware enhancement network is implemented within the PyTorch framework. All training and inference experiments are conducted on a single NVIDIA RTX 3090 GPU [51]. Parameter configurations and training strategies for both stages are detailed comprehensively to guarantee experimental reproducibility. The proposed method adopts a two-stage optimization paradigm characterized by reconstruction followed by enhancement, conducting Zero-Shot instance-level optimization for each individual input image.

In the explicit geometric reconstruction of the first stage, 70,000 2DGS primitives are initialized. A multi-scale pyramid strategy $S = 2$ is adopted to capture geometric details ranging from coarse to fine. The optimization employs the Adam optimizer with an initial learning rate of 0.01. The reconstruction process spans 20,000 iterations, with the parameter $\lambda$ in the loss function $\ell_{rec}$ configured as 0.7.

In the second stage, the Gaussian geometric parameters $\Theta_{frozen}$ learned in the first stage are frozen. Only the parameters of the enhancement network are optimized. The first seven layers of the pre-trained MobileNetV2 serve as the feature extraction backbone, projecting feature dimensions to 32 via a $1 \times 1$ convolution. Simultaneously, the enhancement dictionary is composed of $K = 30$ atoms learned via K-Means clustering and one zero-transformation atom. Consequently, a total of 31 primitive channels are constituted.

During this phase, the enhancement network is trained for a total of 30,000 iterations. The optimization utilizes the Adam optimizer initialized with a learning rate of 0.001. Furthermore, a cosine annealing scheduler is applied to progressively attenuate the learning rate to 5% of its starting magnitude. Based on empirical experiments and grid search, Weight coefficients for each component in the total loss function $\ell_{total}$ are set as follows : $\lambda_1 = 0.01, \lambda_2 = 1, \lambda_3 = 6, \lambda_4 = 0.01, \lambda_5 = 3, \lambda_6 = 0.4$.

Regarding the rendering configuration, a tile-based rasterizer is utilized for 2D Gaussian coefficient rendering with the tile size set to $16 \times 16$. To eliminate artifacts stemming from convolution operations and tile boundaries, reflection padding is applied during both feature extraction and gradient calculation.

**Evaluation Metrics.** Seven widely recognized evaluation metrics are selected in this study for the domain of image restoration and enhancement. These metrics encompass three Full-Reference (FR) indicators and four No-Reference (NR) indicators. Three FR metrics are adopted to assess the consistency between enhanced images and reference images for datasets containing paired Ground Truth (GT). Specifically, Peak Signal-to-Noise Ratio (PSNR), Structural Similarity (SSIM), and Learned Perceptual Image Patch Similarity (LPIPS) are utilized. PSNR quantifies the pixel-level fidelity of the enhanced image. SSIM assesses image similarity across the three dimensions of luminance, contrast, and structure. Meanwhile, LPIPS computes perceptual similarity by measuring the distance between the enhanced and reference images within the feature space of a VGG-based deep neural network.

To evaluate naturalness, contrast, and information content in reference-free scenarios involving unpaired data or real-world applications, this study employs four No-Reference metrics including Natural Image Quality Evaluator (NIQE), Lightness Order Error (LOE), Discrete Entropy (DE), and Enhancement Measure Evaluation (EME). Specifically, NIQE quantifies the degree of naturalness by measuring the distance between the enhanced image and a statistical model of natural images. LOE evaluates the capability to preserve naturalness regarding lightness order. DE gauges the richness of information contained within the image. EME relies on Weber's Law to quantify local contrast variations.

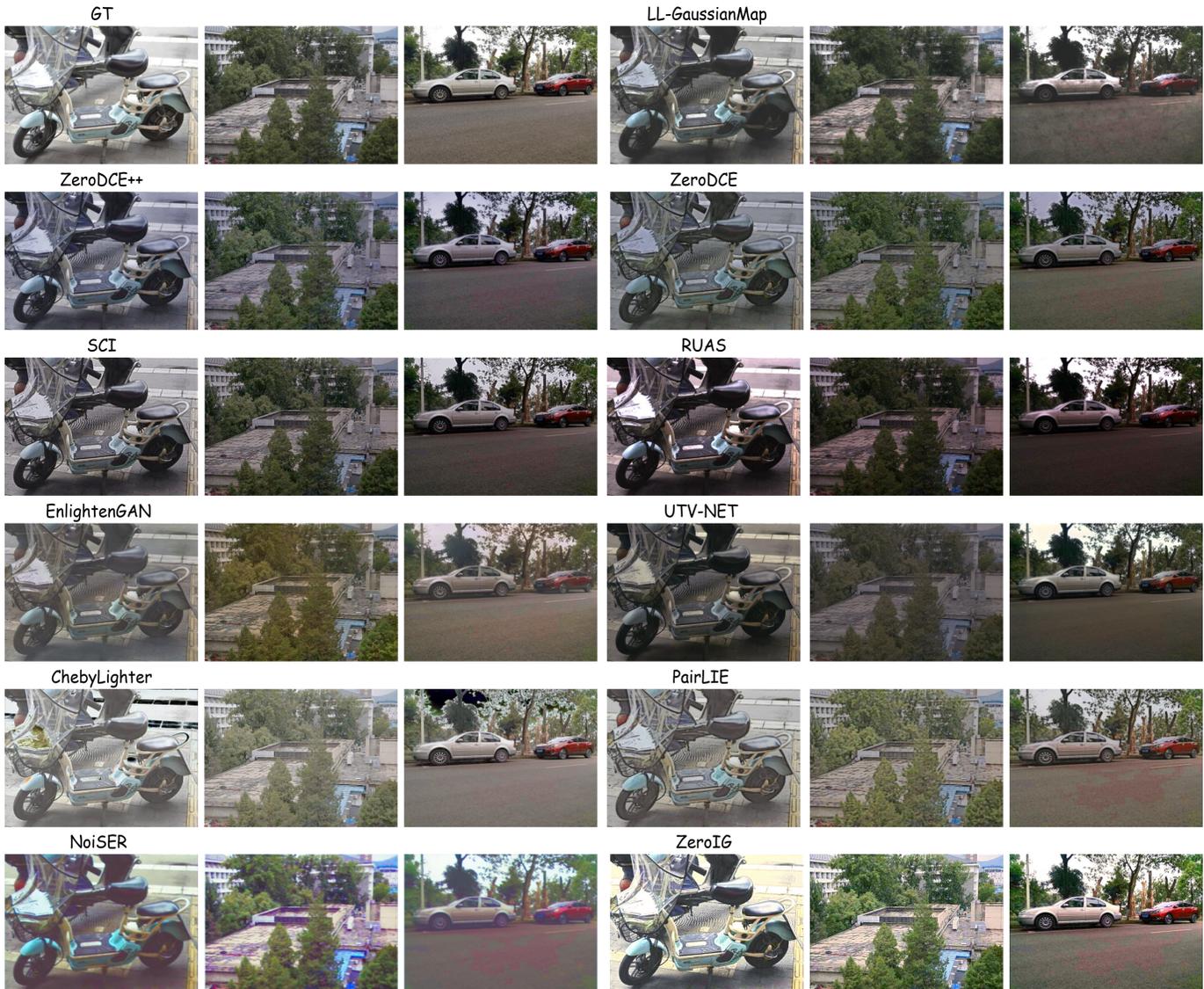

**Fig. 7.** Visual comparison between LL-GaussianMap and SOTA methods on the LOL, LSRW-Huawei, and LSRW-Nikon datasets.

### B. Performance Comparison

To ensure the fairness, ten state-of-the-art (SOTA) unsupervised methods in the LLIE domain are selected for comparison with LL-GaussianMap [7-11,14-15,17-18,20]. Evaluation results on the LOL and LSRW datasets are reported in Tables I and II respectively [49-50].

As illustrated in Fig.7, visual comparison results are reported for images selected from the LOL, LSRW (Huawei), and LSRW (Nikon) datasets. From the perspective of overall brightness, LL-GaussianMap achieves the most faithful brightness restoration among all compared methods. ZeroIG produces images with brightness levels substantially higher than those of the ground truth, whereas methods such as ZeroDCE, EnlightenGAN, and UTV-Net exhibit noticeable under-enhancement. With respect to color reproduction, LL-GaussianMap maintains accurate color consistency without observable color shifts. In contrast, global color deviations are observed in methods including RUAS and EnlightenGAN. In terms of contrast preservation, LL-GaussianMap yields slightly higher contrast, while ZeroDCE, ZeroDCE++, and ChebyLighter tend to produce lower-contrast results, which leads to the loss of fine texture details. Overall, these visual comparisons demonstrate that LL-GaussianMap delivers competitive and well-balanced enhancement performance as a newly developed low-light image enhancement framework. To further evaluate the detail restoration capability of LL-GaussianMap, more challenging scenes from the LSRW dataset are examined in Figs.8 and 9. As shown in Figs.8, LL-GaussianMap achieves accurate color reproduction and preserves rich texture details. Although the overall brightness remains slightly lower than that of the reference image, the contrast is well maintained and fine structural details are clearly retained. In Fig.9, a minor loss of highlight details is observed. Nevertheless, the enhanced result produced by LL-GaussianMap remains highly consistent with the ground truth in terms of overall color appearance and contrast distribution. These results indicate that LL-GaussianMap is capable of delivering stable and visually coherent detail restoration even in complex low-light scenarios.

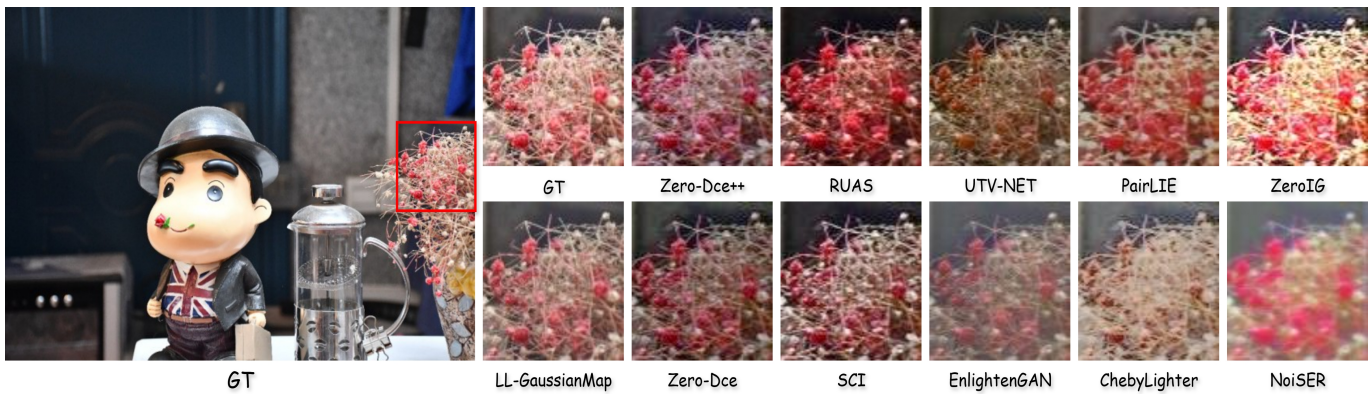

**Fig. 8.** Comparison of detailed features between LL-GaussianMap and SOTA methods on the LSRW dataset. Zoomed-in details within the red box are displayed for each method.

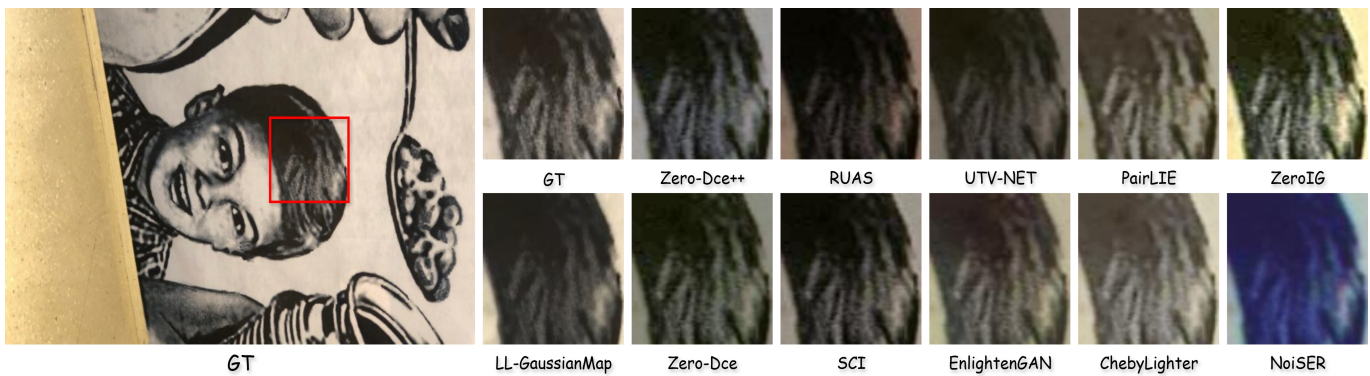

**Fig. 9.** Comparison of detailed features between LL-GaussianMap and SOTA methods on the LSRW dataset. Zoomed-in details within the red box are displayed for each method.

For quantitative evaluation, both Full-Reference (FR) and No-Reference (NR) image quality metrics are employed. Specifically, PSNR, SSIM, and LPIPS are adopted as FR metrics, while NIQE, LOE, DE, and EME are used as NR metrics. As reported in Tables I, II, and III, LL-GaussianMap consistently ranks within the top two in terms of PSNR and SSIM across all four datasets. Moreover, competitive performance is also observed for the remaining evaluation metrics, where the proposed method achieves strong overall rankings.

Overall, although LL-GaussianMap exhibits certain limitations, such as incomplete preservation of fine details in some enhancement cases, it represents the first attempt to introduce 2D Gaussian Splatting scene representation into low-light image enhancement. Considering this novelty, LL-GaussianMap achieves a favorable balance between enhancement quality and methodological innovation, and demonstrates a clear performance advantage when compared with a wide range of state-of-the-art methods.

### C. Ablation Study

To evaluate the effectiveness and contribution of individual components within the proposed framework comprehensively, a series of controlled ablation studies are conducted. All ablation experiments are performed under identical dataset splits and training configurations. Furthermore, the aforementioned evaluation metrics are adopted as constant standards.

**Analysis of Data-Driven Dictionary Construction.** The data-driven manifold dictionary serves as a critical bridge connecting implicit illumination distribution with explicit Gaussian geometry. The sampling density of the model regarding the illumination manifold is determined by the size of the dictionary. Comparative experiments are conducted by setting K to $\{10,30,50,100\}$. As illustrated in Fig. 10, dense weight maps $\mathbf{\Omega}_k$ corresponding to different cluster quantities are displayed. It is evident that detailed textures increase gradually as the value of K rises. Fragmentation in the overall weight distribution emerges when $K>30$. Consequently, unnecessary computational overhead is induced. As presented in TABLE IV, the best performance across all metrics is achieved when $K=30$. Thus, $K=30$ is selected in this paper to balance representation capability and computational efficiency.

**Complexity of Enhancement Curves.** The degrees of freedom of the atom-wise adjustment curve are controlled by P, as defined in (7). This parameter determines the number of learnable coefficients $a_1, a_2...a_p$ assigned to each pixel or dictionary atom for modulating its brightness adjustment curve. Accordingly, $P \in \{1,2,3,5\}$ is evaluated. As shown in Fig. 11, finer texture details are progressively recovered as P increases, and the performance stabilizes when $P = 5$. As reported in TABLE V, the best overall results across all evaluation metrics are achieved at $P = 5$. Therefore, $P = 5$ is adopted as the default parameter setting in this work.

TABLE I
PERFORMANCE COMPARISON RESULTS ON THE LOL DATASET. RED AND BLUE COLOR MARKERS INDICATE THE FIRST AND SECOND PLACE IN PERFORMANCE COMPARISON RESULTS FOR A SINGLE METRIC, RESPECTIVELY.

| Method | SSIM↑ | PSNR↑ | LPIPS↓ | NIQE↓ | LOE↓ | DE↑ | EME↑ |
|---|---|---|---|---|---|---|---|
| ZeroDCE++ [7] | 0.78 | 17.64 | 0.21 | 3.47 | 21.03 | 1.90 | 13.13 |
| ZeroDCE [8] | 0.76 | 17.53 | 0.23 | 3.60 | 29.53 | 1.66 | 13.34 |
| SCI [9] | 0.69 | 17.54 | 0.20 | 3.27 | 8.01 | 1.62 | 14.96 |
| RUAS [10] | 0.52 | 14.40 | 0.24 | 4.11 | 0.60 | 1.08 | 16.20 |
| EnlightenGAN [11] | 0.66 | 15.92 | 0.28 | 3.74 | 124.91 | 1.31 | 2.83 |
| UTV-NET [14] | 0.77 | 17.70 | 0.15 | 4.23 | 20.02 | 1.78 | 6.82 |
| ChebyLighter [15] | 0.69 | 15.64 | 0.30 | 3.51 | 48.21 | 1.96 | 5.39 |
| PairLIE [17] | 0.78 | 19.44 | 0.22 | 4.21 | 58.33 | 1.82 | 5.20 |
| NoiSER [18] | 0.59 | 14.47 | 0.41 | 4.37 | 31.73 | 0.71 | 1.86 |
| ZeroIG [20] | 0.74 | 20.69 | 0.27 | 3.40 | 10.73 | 1.93 | 12.86 |
| LL-GaussianMap | 0.79 | 19.66 | 0.26 | 3.22 | 20.44 | 1.81 | 11.49 |

TABLE II
PERFORMANCE COMPARISON RESULTS ON THE LSRW-HUAWEI DATASET. RED AND BLUE COLOR MARKERS INDICATE THE FIRST AND SECOND PLACE IN PERFORMANCE COMPARISON RESULTS FOR A SINGLE METRIC, RESPECTIVELY.

| Method | SSIM↑ | PSNR↑ | LPIPS↓ | NIQE↓ | LOE↓ | DE↑ | EME↑ |
|---|---|---|---|---|---|---|---|
| ZeroDCE++ [7] | 0.64 | 20.54 | 0.24 | 3.28 | 22.15 | 1.83 | 11.74 |
| ZeroDCE [8] | 0.62 | 18.29 | 0.25 | 3.38 | 43.80 | 1.61 | 11.39 |
| SCI [9] | 0.63 | 19.02 | 0.19 | 3.29 | 11.30 | 1.90 | 14.28 |
| RUAS [10] | 0.52 | 14.35 | 0.22 | 3.08 | 0.38 | 1.50 | 15.04 |
| EnlightenGAN [11] | 0.64 | 19.98 | 0.20 | 2.96 | 50.08 | 1.62 | 7.30 |
| UTV-NET [14] | 0.56 | 16.36 | 0.27 | 3.43 | 12.99 | 1.12 | 6.63 |
| ChebyLighter [15] | 0.62 | 15.72 | 0.23 | 2.43 | 14.28 | 1.81 | 5.18 |
| PairLIE [17] | 0.64 | 18.99 | 0.27 | 5.14 | 77.37 | 1.65 | 6.60 |
| NoiSER [18] | 0.62 | 17.12 | 0.54 | 4.18 | 93.78 | 2.09 | 4.83 |
| ZeroIG [20] | 0.54 | 16.14 | 0.30 | 4.05 | 19.14 | 2.48 | 13.00 |
| LL-GaussianMap | 0.69 | 21.59 | 0.16 | 3.35 | 27.89 | 2.52 | 17.94 |

TABLE III
PERFORMANCE COMPARISON RESULTS ON THE LSRW-NIKON DATASET. RED AND BLUE COLOR MARKERS INDICATE THE FIRST AND SECOND PLACE IN PERFORMANCE COMPARISON RESULTS FOR A SINGLE METRIC, RESPECTIVELY.

| Method | SSIM↑ | PSNR↑ | LPIPS↓ | NIQE↓ | LOE↓ | DE↑ | EME↑ |
|---|---|---|---|---|---|---|---|
| ZeroDCE++ [7] | 0.75 | 13.90 | 0.17 | 4.84 | 123.32 | 1.38 | 12.46 |
| ZeroDCE [8] | 0.73 | 12.80 | 0.18 | 4.92 | 48.91 | 0.84 | 10.95 |
| SCI [9] | 0.73 | 14.19 | 0.14 | 5.18 | 32.33 | 1.37 | 14.93 |
| RUAS [10] | 0.64 | 11.35 | 0.15 | 4.99 | 0.24 | 1.43 | 16.37 |
| EnlightenGAN [11] | 0.79 | 18.29 | 0.14 | 4.85 | 98.16 | 1.51 | 4.32 |
| UTV-NET [14] | 0.68 | 11.05 | 0.19 | 5.25 | 20.97 | 0.90 | 5.57 |
| ChebyLighter [15] | 0.79 | 20.14 | 0.14 | 5.06 | 47.19 | 1.48 | 5.16 |
| PairLIE [17] | 0.77 | 16.35 | 0.18 | 5.85 | 89.78 | 1.10 | 5.34 |
| NoiSER [18] | 0.74 | 16.68 | 0.35 | 6.20 | 25.13 | 1.70 | 4.30 |
| ZeroIG [20] | 0.74 | 17.38 | 0.16 | 4.63 | 65.08 | 1.70 | 10.94 |
| LL-GaussianMap | 0.86 | 19.06 | 0.19 | 4.97 | 24.12 | 1.81 | 15.14 |

**Contribution of Loss Terms.** The overall loss function $\ell_{total}$ incorporates exposure control, spatial consistency, sparsity regularization, and perceptual quality. To assess the contribution of each component, ablation experiments are conducted by removing individual loss terms. The corresponding quantitative visual comparisons are summarized in Fig.12. The results indicate that each loss component plays a critical role in the enhancement process. In particular, removing the exposure control loss $\ell_{exp}$ and the spatial consistency loss $\ell_{cont}$ leads to noticeable brightness degradation. When the total variation loss $\ell_{tv}$ is excluded, pronounced color transition artifacts and an imbalance in global contrast are observed.

**Impact of Enhancement Iterations.** A zero-shot learning strategy is adopted, in which performance gains are obtained through iterative optimization of illumination enhancement parameters on a single image. The iteration number directly affects optimization convergence and visual quality. To analyze this effect, the model is evaluated systematically under different iteration settings, including 5K, 10K, 20K, 50K, and 100K. As reported in TABLE VI, both brightness restoration and detail clarity improve progressively as the iteration count increases, reaching a maximum at 50K iterations, after which a gradual degradation is observed. The results indicate that 50K iterations achieve the best balance between global brightness recovery and color fidelity. Further analysis of TABLE VI shows that this setting yields the best performance on most evaluation metrics. With the exception of LOE and EME, all remaining metrics attain their highest rankings at 50K iterations. Based on these experimental findings, 50,000 iterations are adopted as the default configuration for the second-stage optimization.

TABLE IV
PERFORMANCE EVALUATION RESULTS OF LL-GAUSSIANMAP UNDER DIFFERENT K VALUES. RED AND BLUE COLOR MARKERS INDICATE THE FIRST AND SECOND PLACE IN PERFORMANCE COMPARISON RESULTS FOR A SINGLE METRIC, RESPECTIVELY.

| K | SSIM↑ | PSNR↑ | LPIPS↓ |
|---|---|---|---|
| 10 | 0.76 | 17.58 | 0.24 |
| 30 | 0.79 | 19.55 | 0.21 |
| 50 | 0.73 | 17.48 | 0.24 |
| 100 | 0.72 | 17.32 | 0.23 |

TABLE V
PERFORMANCE EVALUATION RESULTS OF LL-GAUSSIANMAP UNDER DIFFERENT P VALUES. RED AND BLUE COLOR MARKERS INDICATE THE FIRST AND SECOND PLACE IN PERFORMANCE COMPARISON RESULTS FOR A SINGLE METRIC, RESPECTIVELY.

| P | SSIM↑ | PSNR↑ | LPIPS↓ |
|---|---|---|---|
| 1 | 0.62 | 16.22 | 0.43 |
| 2 | 0.71 | 17.25 | 0.29 |
| 3 | 0.73 | 17.23 | 0.23 |
| 5 | 0.79 | 19.55 | 0.21 |

TABLE VI
PERFORMANCE EVALUATION RESULTS OF LL-GAUSSIANMAP UNDER DIFFERENT ITERATION COUNTS. RED AND BLUE COLOR MARKERS INDICATE THE FIRST AND SECOND PLACE IN PERFORMANCE COMPARISON RESULTS FOR A SINGLE METRIC, RESPECTIVELY.

| Enhance Iterations | SSIM↑ | PSNR↑ | LPIPS↓ |
|---|---|---|---|
| 5K | 0.64 | 16.74 | 0.33 |
| 10K | 0.72 | 17.55 | 0.24 |
| 20K | 0.76 | 18.48 | 0.23 |
| 50K | 0.79 | 19.55 | 0.21 |
| 100K | 0.75 | 18.52 | 0.24 |

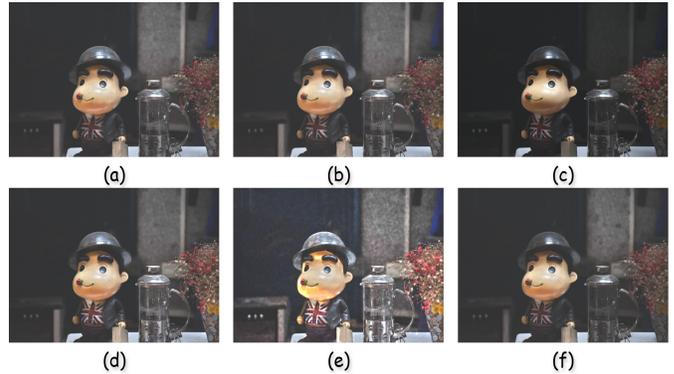

Fig. 12. Visual comparison of the impact of different loss functions on enhancement results. (a)-(f) represent the final visualization results with specific losses removed, respectively

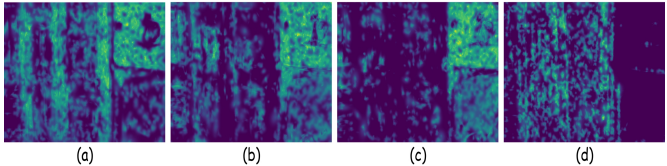

Fig. 10. Visualization of dense weight $\mathbf{\Omega}_k$ maps corresponding to different cluster quantities, where, (a)-(d) represent the visualization effects for K ∈ {10,30,50,100} respectively.

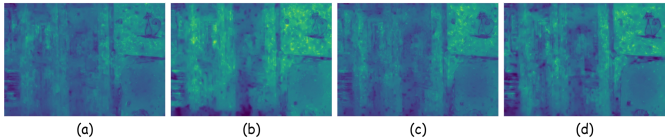

Fig. 11. Visualization of spatial gain maps η(x) corresponding to different P, where, (a)-(d) represent the visualization effects for P ∈ {1,2,3,5} respectively.

*D. Limitation and Future Works*

Remarkable generalization capabilities and high-fidelity visual effects have been demonstrated by LL-GaussianMap in low-light enhancement tasks. Nevertheless, several avenues warranting further exploration remain within this domain. First, achieving a better balance between inference efficiency and reconstruction quality remains an open challenge. Achieving real-time inference without degrading perceptual quality has become an important research focus, motivated by recent advances in feed-forward Gaussian splatting techniques. The currently adopted two-stage optimization paradigm ensures instance-level adaptability for individual images, but introduces notable computational overhead compared with pure inference-based models. Future work may explore end-to-end network architectures to accelerate the prediction of Gaussian geometric parameters and enhancement coefficients, thereby enabling more efficient deployment. Second, reconstructing intricate high-frequency texture details using discrete Gaussian ellipsoids remains an open challenge. Detail blurring may occur when extremely fine textures are approximated with a limited number of primitives, which stems from the inherent smoothing property of Gaussian kernels. More expressive geometric primitives will be investigated in future work, or frequency-domain constraints will be incorporated to further improve high-frequency reconstruction accuracy. Finally, LL-GaussianMap is expected to facilitate the advancement of low-level vision applications within the compressed 2DGS image representation domain. Images are encoded as compact and structured sets of 2D Gaussian primitives, which may enable next-generation image processing algorithms that jointly achieve storage efficiency and structural awareness.

V. CONCLUSION

LL-GaussianMap is presented as a pioneering Zero-Shot unsupervised framework that introduces 2D Gaussian Splatting into low-light image enhancement in this paper. This framework addresses the inherent limitations of implicit representations in structure preservation and artifact suppression. The enhancement task is reformulated as a gain map generation problem guided by high-fidelity geometric priors. A data-driven manifold dictionary and a unified enhancement module are jointly employed, enabling deep integration of illumination modeling and explicit geometric representation through differentiable Gaussian rasterization. As a result, image edges are preserved effectively and natural brightness restoration is achieved without reliance on paired training data. Limitations in real-time inference efficiency remain due to the current scene-wise optimization strategy. Moreover, geometric robustness under extreme noise conditions requires further investigation. Nevertheless, extensive experimental results validate the strong potential of explicit Gaussian representations in low-level vision tasks. Future work will focus on developing feed-forward Gaussian encoders to replace iterative optimization, as well as extending the proposed framework to joint denoising and other image restoration tasks. These efforts are expected to open new directions for image processing paradigms based on explicit primitives.

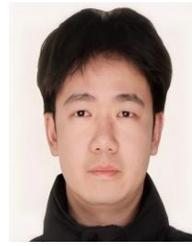

**Yuhan Chen** received his master's degree in 2024 from the College of Mechanical Engineering at Chongqing University of Technology. He is currently pursuing the Ph.D. degree in College of Mechanical and Vehicle Engineering at Chongqing University, China. His research interests include deep learning, Low-level Vision and Gaussian Splatting.

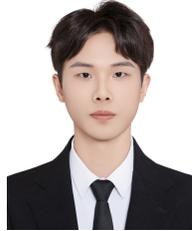

**Ying Fang** received the B.E. degree majoring in Vehicle Engineering at Chongqing University of Technology in 2024. He is currently pursuing the M.S. degree in Mechanical Engineering at Chongqing University, Chongqing, China. His research interests include computer vision, Gaussian Splatting and deep learning.

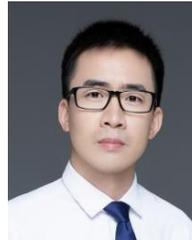

**Guofa Li** received the Ph.D. degree in Mechanical Engineering from Tsinghua University, China, in 2016. He is currently a Professor with Chongqing University, China. His research interests include environment perception, driver behavior analysis, and smart decision-making based on artificial intelligence technologies in autonomous vehicles and intelligent transportation systems. He serves as the Associate Editor for *IEEE Transactions on Intelligent Transportation Systems, IEEE Transactions on Affective Computing,* and *IEEE Sensors Journal.*

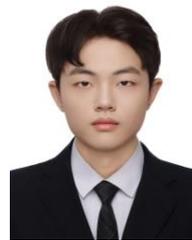

**Wenxuan Yu** received the B.E degree majoring in Mechanical Design, Manufacturing, and Automation at Chongqing University in 2025. He is currently pursuing the M.S. degree in Mechanical Engineering at Chongqing University, Chongqing, China. His research interests include computer vision, Gaussian Splatting and deep learning.

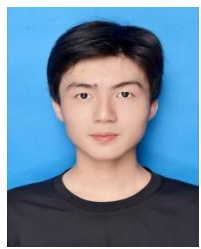

**Yicui Shi** received the B.E degree majoring in Automotive Engineering at Chongqing University in 2025. He is currently pursuing the M.S. degree in Automotive Engineering at Chongqing University, Chongqing, China. His research interests include computer vision and Gaussian Splatting.

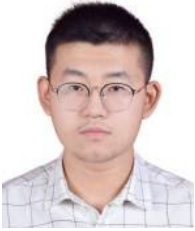
**Jingrui Zhang** received the B.E. degree from Xiamen University in 2024. He is currently pursuing the M.S. degree in Software Engineering at the School of Computer Science, Wuhan University. His research interests include image processing and machine vision.

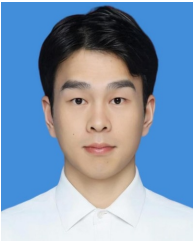
**Kefei Qian** received his master's degree in 2024 from the College of Mechanical and Vehicle Engineering at Chongqing University. He is currently pursuing the PhD. degree in College of Mechanical and Vehicle Engineering at Chongqing University, China. His research interests include 3D/4D reconstruction, sensor simulation and generative models.

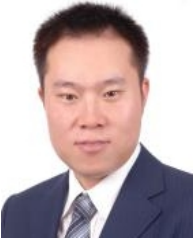
**Wenbo Chu** received his B.S. degree majored in Automotive Engineering from Tsinghua University, China, in 2008, and his M.S. degree majored in Automotive Engineering from RWTH-Aachen, German and Ph.D. degree majored in Mechanical Engineering from Tsinghua University, China, in 2014. He is currently a research fellow at Western China Science City Innovation Center of Intelligent and Connected Vehicles (Chongqing) Co, Ltd., and National Innovation Center of Intelligent and Connected Vehicles.

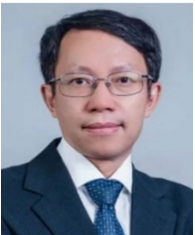
**Keqiang Li** received the B.E. degree from Tsinghua University, Beijing, China, in 1985, and the M.E. and Ph.D. degrees from Chongqing University, Chongqing, China, in 1988 and 1995, respectively. He is currently a Professor with the School of Vehicle and Mobility, Tsinghua University. He is the Chief Scientist of Intelligent and Connected Vehicle Innovation Center of China, and the Director of State Key Laboratory of Automotive Safety and Energy of China. His current research interests include intelligent connected vehicles, cloud-based control for vehicles, and vehicle dynamics systems.